\documentclass{article}
\usepackage{spconf,amsmath,graphicx,booktabs,chemformula,enumitem}

\title{A-CCNN: Adaptive CCNN for Density Estimation and Crowd Counting}
\name{Saeed Amirgholipour Kasmani$^1$, Xiangjian He$^{1,}$\sthanks{Corresponding author.}, Wenjing Jia$^1$, Dadong Wang$^2$, Michelle Zeibots$^3$}
\address{$^1$ Global Big Data Technologies Centre,  University of Technology Sydney, 
	Australia\\
$^2$ Quantitative Imaging, CSIRO Data61,  Australia\\
$^3$Institute for Sustainable Futures, University of Technology Sydney, Australia}

\begin{document}

\maketitle
\begin{abstract}

Crowd counting, for estimating the number of people in a crowd using vision-based computer techniques, has attracted much interest in the research community. Although many attempts have been reported, real-world problems, such as huge variation in subjects' sizes in images and serious occlusion among people, make it still a challenging problem. In this paper, we propose an Adaptive Counting Convolutional Neural Network (A-CCNN) and consider the scale variation of objects in a frame adaptively so as to improve the accuracy of counting. Our method takes advantages of contextual information to provide more accurate and adaptive density maps and crowd counting in a scene. 
Extensively experimental evaluation is conducted using different benchmark datasets for object-counting and shows that the proposed approach is effective and outperforms state-of-the-art approaches.
\end{abstract}

\section{Introduction}
\label{sec:intro}

Nowadays, density estimation and counting the number of people in a crowded scene is a desirable application especially in restricted, public event places such as train stations. 
Incidents, traffic delay and even terrible stampedes may be caused by overcrowding in such a scene. 
Generally, there is an urgent need for real-time decision making corresponding to crowd changes. To deal with this situation, there exist various challenges caused by occlusions, size and shape variations of people, perspective distortion, etc. Thus, correctly counting in crowded areas is very necessary in many real world applications including visual surveillance, traffic monitoring and crowd analysis.

The existing approaches for crowd density estimation can be divided into two main groups, i.e., detection based methods and feature regression based methods~\cite{sindagi2017survey}. 
Detection based methods (also called direct methods) segment and detect every individual people or objects in a scene with pre-trained classifiers and then simply count them. 
However, in  complex scenes with serious occlusions and extremely crowded scenes, these approaches often fail to detect individuals and therefore produce inaccurate countings. 
In the feature regression based approaches (also called indirect approaches), learning algorithms or statistical methods are utilized to analyze the image appearance features of a crowded scene, and then estimate the number of people or objects based on image appearance. Thus, these methods are more suitable for dealing with highly crowded scenes where detecting individuals often fails.

\begin{figure}[t]
  \includegraphics[width=8.5cm]{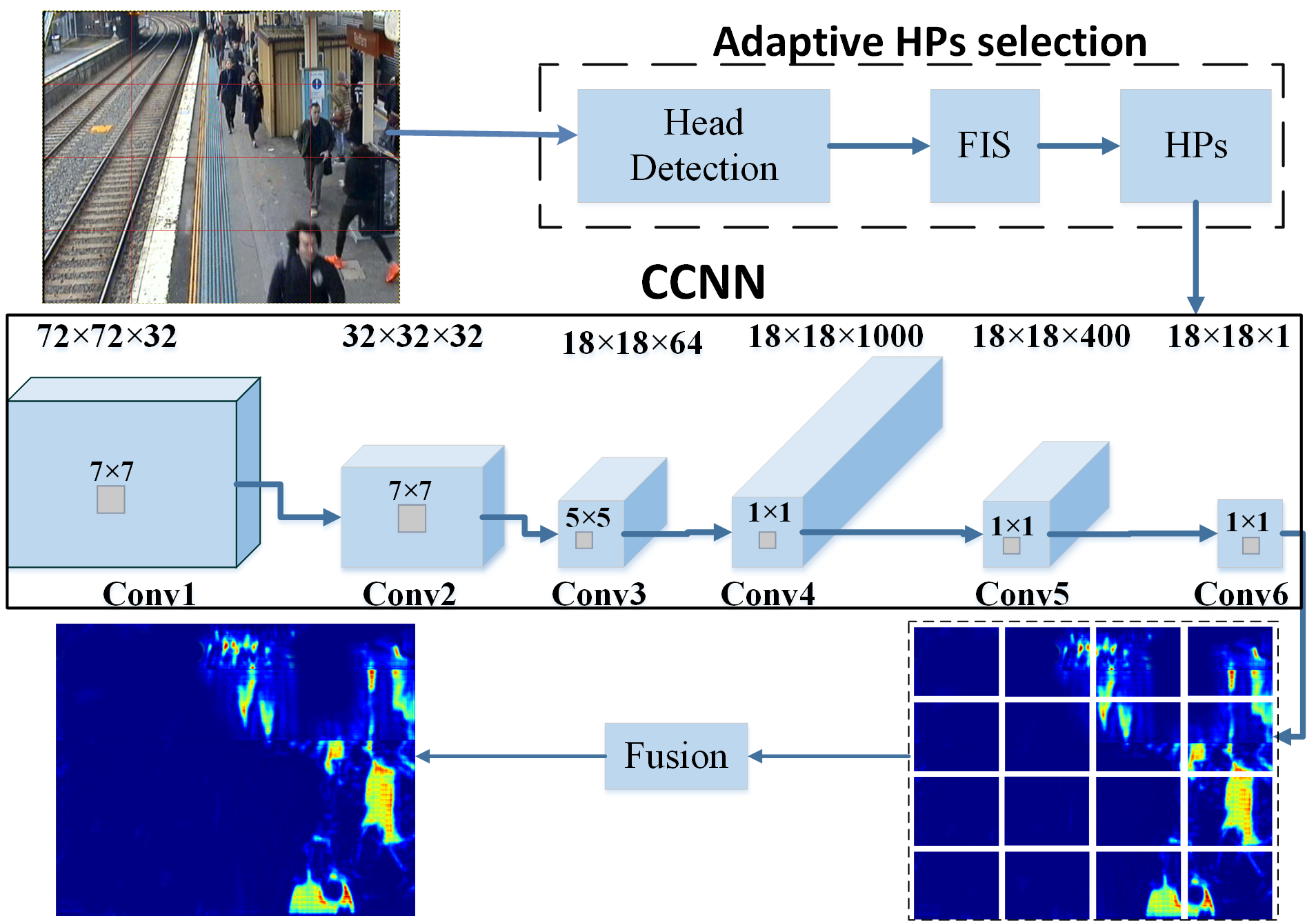}
  \caption{The overview of our proposed A-CCNN crowd counting method. For an input image, our A-CCNN first estimates head size and corresponding position, and then utilizes a fuzzy engine to determine HP of the CCNN models for generating the density map.}
  \label{fig:1}
\end{figure}

In this paper, based on the recent advance of Counting Convolutional Neural Network (CCNN)~\cite{onoro2016towards}, we propose a new adaptive CCNN architecture, abbreviated as A-CCNN, that processes each part of an input image using an optimally trained CCNN model in order to estimate the corresponding density map accurately. 
As illustrated in Fig.~\ref{fig:1}, to tackle the counting problem, our A-CCNN model is able to regress the density function corresponding to a specified section. This allows our model to accurately localize density maps for unseen images.

The most noticeable properties that make the proposed model outstanding for crowd analysis are: (1) the ability to handle large scale variations in people's sizes when appearing in images; and (2) the facility to generate local density maps within a crowd scene. Therefore, the proposed model can give a complete view about the scattering of a crowd. 
Compared to the prior works, our approach does not use different CCNN architectures, and only tries to select the most effective Hyper Parameters (HPs)  for generating a CCNN model. Thus, it can learn to address scale variations in an image with a simple and effective way. 
\section{Related Works}
\label{sec:reworks}

In recent years, many researchers~\cite{zeng2017multi, sindagi2017cnn,sam2017switching} have developed deep learning models for image segmentation, classification and recognition, and achieved very good results. 
Inspired by these, Convolutional Neural Network (CNN) models have been proposed to learn to count people and produce density maps in images simultaneously, and they have worked well for objects of approximately the same size in an image or a video.
Sindagi and Patel~\cite{shang2016end} proposed an end-to-end cascaded network of CNNs that can learn globally relevant and discriminative features to estimate highly refined density maps with low count errors. 
Onoro-Rubio and Lopez-Sastre proposed a regression model called Counting CNN (CCNN)~\cite{onoro2016towards}, which can map the appearances of input image patches to corresponding density maps. They also proposed a Hydra CNN based on the idea of multi-scaling crowd counting and achieved a sufficient advantages in comparison with the previous models.

Inspired by the Hydra CNN method, some researchers have tried to utilize more complex deep models to solve the problem caused by the significant variance of crowd's appearance in a captured image/video. 
Deepak et al.~\cite{sam2017switching} proposed a switching CNN to select the best CNN regressor for each of different receptive fields and achieved better results than the state-of-the-arts for crowd counting.
Kumagai1 et al.~\cite{kumagai2017mixture} proposed a mixture of CCNNs and adaptively selected multiple CNNs according to the appearance of a test image for predicting the number of people. 
Zhang et al. proposed a multi-column network and three independent CNN architectures, and then used the combined features of these three networks to get a density map~\cite{zeng2017multi}.

Our work presented in this paper is based on the CCNN architecture~\cite{onoro2016towards}. 
The CCNN approach takes a small patch of the input image as input and generates the corresponding density map for the image patch. By utilizing the sliding window technique, it extracts patches and applies a CNN model to regress the density function. 
Therefore, CCNN is formulated as a regression model that generates object density maps based on the corresponding appearances of image patches. 

Formally, in the original CCNN model, the ground truth density map \ch{D_I} is defined as,
\begin{equation}
D_{I}(p)=\sum_{\mu \epsilon A_{I}} N(p,\mu,\Sigma),
\label{eq:1}
\end{equation}
where $A_I$ represents the number of annotated points in the image $I$, and $N$ ($p$; $\mu$; $\Sigma$ ) represents a normalized 2D Gaussian function with a mean of $\mu$ and a covariance of $\Sigma$, evaluated at each pixel position $p$.

The CCNN utilizes two important HPs for generating models, i.e., the patch size and the value of $\Sigma$ in the Gaussian function.
Through careful analysis of CCNN, we have noticed that it has a major problem in producing a correct density map, because CCNN treats the whole parts of the input image in the same way. 
Therefore, CCNN cannot achieve an acceptable accuracy in density estimation when a scene has a large scale variation in the sizes of objects. 
We have observed that more accurate density maps can be produced when the values of the above mentioned two HPs are optimally and properly chosen.

\section{Adaptive CCNN}
\label{sec:method}

In our work, in order to handle crowd images with large varieties in targets' appearances, we propose a new A-CCNN model for crowd counting. 
As shown in Fig.~\ref{fig:1}, our A-CCNN architecture takes an image as input and equally divides the image to 16 parts and then determines the average of  heads' sizes and position in different parts of an image. Then, by utilizing a Fuzzy Inference System (FIS), it feeds each image section with the same FIS linguistic output value to an appropriate CCNN model with a proper HP to obtain the corresponding density map for each section. In the end, it merges the output of different parts to obtain the final density map output. 

In reality, the sizes of people who are closer to the camera appear to be bigger than those of the people who are further from the camera. Based on our observation from expriments, we find that there is a relationship between HPs and the scales of people. Thus, we use smaller (and larger) $\Sigma$'s and patch sizes for the areas containing smaller (and larger) targets. 
Then, we train CCNN models to create density maps for different sizes of patches. For each image in the testing stage, our A-CCNP model extracts image patches from it, and generates their corresponding object density maps by utilizing the relative CCNN models according to their sizes. Then, the density maps of these patches are assembled into the density map of the testing image.

Compared with CCNN, we have made the following improvement in the proposed A-CCNN. 
Firstly, we use different patch sizes in A-CCNN according to the sizes of people in the patches, different from using the same patch size for all patches in the original CCNN. Secondly, we have utilized different $\Sigma$ values to generate the training patches. The $\Sigma$ of the Gaussian function in Eq.~\ref{eq:1} is changed to adapt the size of a patch.
In comparison with the Switch-CNN, our proposed A-CCNN uses only one well-known CCNN model with adaptive HPs, so it has less complexity than the Switch-CNN with different CNN architectures.

The process of the proposed A-CCNN is summarized as follows and detailed in the following subsections. First, we perform tiny-face detection~\cite{hu2017finding} to estimate the sizes of heads in each patch of an image. 
Then, by feeding the head sizes and the corresponding head positions to a fuzzy inference system (FIS), we generate the appropriate HPs corresponding to the patches. 
Finally, these HPs are used to train CCNNs that can adaptively generate the density maps for various patches. 

\subsection{Head Detection}
\label{ssec:head}

To obtain the most suitable values of HPs, we need to know the sizes of people or objects in different parts of an image. Therefore, the tiny-face detection approach~\cite{hu2017finding} is used to detect faces in each part of the input image. It creates a coarse image pyramid of the input image, and then feeds the scaled inputs into a CNN to get the template responses. Finally, the final detection results are produced by applying the non-maximum suppression (NMS) at the original resolution.

\subsection{Adaptive HP Selection by FIS}
\label{ssec:fis}

\begin{figure}[h]
  \includegraphics[width=8.5cm]{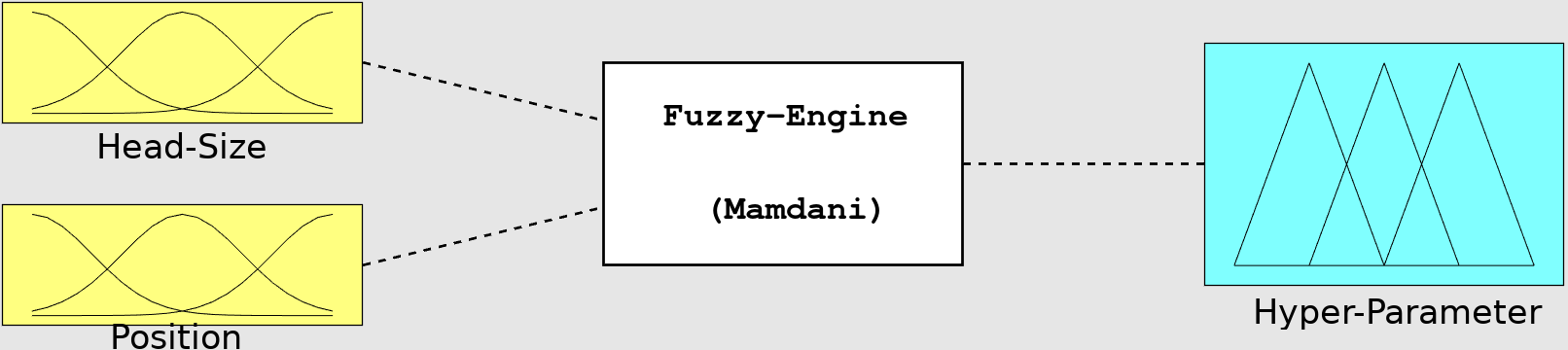}
  \caption{The fuzzy inference engine, where the head size and corresponding position are the two inputs and the level of HPs for CCNN is the output.}
  \label{fig:2}
\end{figure}

As shown in the Fig.~\ref{fig:1}, in order to obtain the values of the HPs, an FIS is designed to adaptively select the values of  HPs according to the sizes and the positions of heads. As shown in Fig.~\ref{fig:2}, the FIS receives the fuzzy information about head sizes and positions, and outputs the fuzzy linguistic variables in the form of fuzzy. 
We choose the same Gaussian membership function for all input and output variables. Small, Average and Big are the fuzzy linguistic variables according to head sizes, and  Up, Middle and Down are the fuzzy linguistic values according to head positions. The output linguistic variables are High-Pred, Mid-Pred, and Low-Pred. 

Based on the Gaussian membership function, the input values are converted into fuzzy linguistic variable in FIS. 
Then, the fuzzy if-then rules developed based on the Mamdani method~\cite{mamdani1976application} are used to map the input variables to appropriate fuzzy output variables. 
In total, nine fuzzy if-then rules are presented in Table ~\ref{table:1}. 
In general, higher (and lower) values of $\Sigma$ and sliding window (patch size) produce density maps with lower (higher) counts of numbers of people.
As an illustration, if an output of FIS is High-Pred (Mid-Pred, Low-Pred), the corresponding CCNN is trained with low (medium, big) HP values.

\begin{table}[h]
  \centering
  \caption{The fuzzy rule table for selecting HPs}
  \begin{tabular}{@{}ccccc@{}}
    \toprule
      & Input &  &  & Output \\
    \cline{1-4}
     Head Size  &  & Position &  &  \\
     \midrule
    Small & & Up &  & High-Pred \\
    Small & & Middle &  & High-Pred \\
    Small & & Down &  & Mid-Pred \\
    Average & & Middle &  & Mid-Pred \\
    Average & & Down &  & Mid-Pred \\
    Average & & Up &  & Low-Pred \\
    Big & & Up &  & Mid-Pred \\
    Big & & Down &  & Low-Pred \\
    Big & & Middle &  & Low-Pred \\
    \bottomrule
  \end{tabular}
  \label{table:1}
  
\end{table}

\subsection{Training Parameters}
\label{ssec:train}

Showing the effectiveness of the proposed A-CCNN, we use the same training parameters as in~\cite{onoro2016towards}, except for two HPs, which are the patch sizes and $\Sigma$'s, for people counting and density estimation.
These two HPs are empiracally determined on the traing data set.
 Similar to the approach in~\cite{onoro2016towards}, a stochastic gradient decent algorithm is used during training.
The momentum, the learning rate and the weight decay are set to be 0.9, 0.0001 and 0.001 respectively. 
After 25 epochs, the model can reach a local optimum.

\section{Experimental Results}
\label{sec:exprim}

To evaluate the performance of our A-CCNN algorithm, experiments are conducted on three challenging crowd counting datasets, i.e., the UCSD dataset~\cite{chan2008privacy}, the UCF-CC dataset~\cite{idrees2013multi}, and the dataset of Sydney Trains Footage (STF)~\cite{farhood2017counting}. Note that the first two are public benchmark datasets.

The Mean Absolute Error (MAE) is used as the evaluation metric for comparing the performance of A-CCNN against the state-of-the-art methods, and it is defined as:
\begin{equation}
MAE=\dfrac{1}{N}\sum_{i=1}^{N} \left | C_{i}- C_{i}^{GT} \right |,
\label{eq:2}
\end{equation}
where $N$ is the number of images, $C_{i}$ is the crowd count predicted by the model being evaluated, and $C_{i}^{GT}$ is the crowd count from the human annotated one (i.e., ground truth).

\subsection{The UCSD Dataset}
\label{ssec:ucsd}

The UCSD crowd counting dataset consists of 2000 frames of size 238$\times$158 from a single far distance scene. We split the dataset into four subsets of training and testing images in the same way as in~\cite{onoro2016towards}. 

Table~\ref{table:2} presents the MAE results for our proposed A-CCNN and six state-of-the-art methods. 
As shown in Table~\ref{table:2}, our A-CCNN performs competitively against other approaches with the lowest ever MAE of $\textbf{1.04}$ and $\textbf{1.48}$ for the upscale and minimal subsets respectively. Furthermore, in the other subsets, the results indicate that A-CCNN outperforms the CCNN by more than $\textbf{8}$ percent. Overall, A-CCNN reaches the best ever average result with MAE of $\textbf{1.35}$.

\begin{table}[h]
  \centering
  \caption{Comparison of the MAE results results between A-CCNN and state-of-the-art crowd counting on UCSD crowd-counting dataset~\cite{chan2008privacy}}
  \begin{tabular}{@{}ccccc|c@{}}
    \toprule
    Methods  & Max & Down & Up & Min & Avg \\
    \midrule
Density Learning~\cite{lempitsky2010learning} & 1.70 & 1.28 & 1.59 & 2.02  & 1.64\\
    Count Forest~\cite{pham2015count} & 1.43 & 1.30 & 1.59 & 1.62 & 1.49 \\
    Arteta et al.~\cite{arteta2014interactive} & \textbf{1.24} & 1.31 & 1.69 & 1.49 & 1.43 \\
    Zhang et al.~\cite{zhang2015cross} & 1.70 & \textbf{1.26} & 1.59 & 1.52 & 1.52 \\
    Switch-CNN~\cite{sam2017switching} & - & - & - & 1.62 & 1.62 \\
    CCNN~\cite{onoro2016towards} & 1.65 & 1.79 & 1.11 & 1.50 & 1.51 \\
    \midrule
    \textbf{\ch{A-CCNN} }& 1.51 & 1.36 & \textbf{1.04} & \textbf{1.48} & \textbf{1.35} \\
    
    \bottomrule
  \end{tabular}
  \label{table:2}
\end{table}

\subsection{The UCF-CC Dataset}
\label{ssec:ucf}

The UCF CC 50~\cite{idrees2013multi} is a small dataset with 50 picture collections of annotated crowd scenes. We have followed the same experimental settings as those of six other state-of-the-art models \cite{sam2017switching}. 

In Table ~\ref{table:3}, the MAE performance of our A-CCNN compared with other methods is shown. As shown in Table ~\ref{table:3}, A-CCNN outperforms four out of six methods and improves the MAE score by more than $\textbf{24}$ percentage compared to the original CCNN. Considering its simplicity, A-CCNN's performance is comparable to that of Switch-CNN and Hydra-CCNN. 

\begin{table}[h]
  \centering
  \caption{Comparison of the MAE results between A-CCNN and state-of-the-art crowd counting on UCF CC dataset ~\cite{idrees2013multi}.}
  \begin{tabular}{@{}ccccccc@{}}
    \toprule
    Methods   &  &  & &  &  &  MAE     \\
    \midrule
    Density learning ~\cite{lempitsky2010learning} &  &  & &  &  &  493.4  \\
    Idrees et al.~\cite{idrees2013multi}  &  &  & &  &  &   419.5 \\
    Zhang et al.~\cite{zhang2015cross}  &  &  & &  &  &   467.0  \\
    MCNN~\cite{zhang2016single} &  &  & &  &  &   377.6  \\
    Hydra-CCNN~\cite{onoro2016towards}  &  &  & &  &  &   333.73 \\
    Switch-CNN~\cite{sam2017switching}  &  &  & &  &  &   \textbf{318.1}  \\
    CCNN~\cite{onoro2016towards}  &  &  & &  &  &   488.67 \\
        \midrule

   \textbf{ \ch{A-CCNN}} &  &  & &  &  &   {367.3} \\
    
    \bottomrule
  \end{tabular}
  \label{table:3}
\end{table}

\subsection{The Sydney Train Footage}
\label{ssec:our}

To evaluate the robustness of our model on real-world problems with heavy occlusions, low resolution and large variance in people's sizes, we have utilized CCTV footages of a train station in Sydney and created annotated data for training and testing with our proposed approach. An example is shown in Fig.~\ref{fig:1}. This dataset has two separate scenes, taken by cameras C5 and C9 with 788 and 600 frames, respectively, with crowd varying between 3 to 65. The sizes of the input frames are 576$\times$704, and the mask and annotation are provided. The huge variation in people's sizes and heavy extreme occlusions make it a very challenging task. 
Generally, in this dataset, the sizes of people who are in front of the cameras are three to four times larger than the sizes of people in further areas. 

Table ~\ref{table:4} reports the MAE performance  on this dataset. 
The crowd count of A-CCNN is significantly higher than the original CCNN. This reinforces the fact that utilizing our A-CCNN can efficiently manage both the difference in appearances and sizes of people. Thus, the various trained CCNNs employed by A-CCNN are able to provide precise density maps, independent of the datasets.

\begin{table}[h]
  \centering
  \caption{Comparison of the MAE results results between A-CCNN and state-of-the-art crowd counting on 
  STF~\cite{farhood2017counting}.}
  \begin{tabular}{@{}cccccccccc@{}}
    \toprule
    Methods   & & & & C5     & &&  & C9 \\
    \midrule
    
    Farhood et al.~\cite{farhood2017counting} & &&  &  2.28  & &&  & 2.67  \\
  CCNN~\cite{onoro2016towards}  & &&  &  3.90  & &&  &  4.23  \\ 
    \midrule
   \textbf{ \ch{A-CCNN}} & &&  &  \textbf{1.69} & &&  &  \textbf{1.87} \\
    \bottomrule
  \end{tabular}
  \label{table:4}
\end{table}

\section{CONCLUSION}
\label{sec:concl}

Aiming to tackle the difficult problem of crowd counting such as scale variance and extreme collusion, we have presented an Adaptive CCNN architecture that takes a whole image as input and directly outputs its density map. 
The proposed method has made a full use of contextual information to generate an accurate density map. 
To leverage the local information, we have utilized the combination of CNN-based head detection and fuzzy inference engine to choose an optimal CCNN model adaptively to each patch of the input image. 
We have achieved noticeable improvements on three challenging datasets, i.e., the UCSD, UCF-CC and the crowd dataset collected by ourselves from a train station in Sydney, and have demonstrated the effectiveness of the proposed approach.

\section{ ACKNOWLEDGMENT}
\label{sec:ack}

        This  work  was  partly  supported  by Rail Manufacturing CRC and Sydney  Trains  with  UTS  project ID PRO17-3968.

\bibliographystyle{IEEEbib}
\bibliography{refs,strings}

\end{document}